\documentclass[letterpaper, 10 pt, conference]{ieeeconf}  

\IEEEoverridecommandlockouts                              
\usepackage{amsfonts}
\usepackage{mathtools}
\usepackage{amsmath}
\usepackage{amssymb}
\usepackage{framed}
\usepackage{balance}
\usepackage{subfigure}
\usepackage{graphics} 
\usepackage{lipsum}
\usepackage{graphicx}
\usepackage{mathptmx} 
\usepackage{algorithm}
\usepackage{amssymb}  
\usepackage{algpseudocode}
\usepackage{booktabs}
\usepackage{tabularx}
\usepackage{booktabs}
\usepackage{array}
\usepackage{etex}
\usepackage{adjustbox}
\usepackage{adjustbox,lipsum}
\usepackage[table]{xcolor}
\usepackage{mdwtab}
\usepackage{gensymb}
\usepackage{nicefrac}
\usepackage{multirow}
\usepackage{colortbl}
\usepackage{hyperref}
\usepackage{color,soul}
\usepackage{epsfig} 
\usepackage{graphics} 

\algrenewcommand\algorithmicforall{\textbf{foreach}}
\algrenewcommand\algorithmicindent{.8em}

\DeclareMathOperator*{\argmax}{arg\,max}  
\DeclareMathOperator*{\argmin}{arg\,min} 

\title{\LARGE \bf
An Architecture for Person-Following using Active Target Search
}

\author{Minkyu Kim$^{1}$, Miguel Arduengo$^{1}$, Nick Walker$^{2}$, Yuqian Jiang$^{2}$, Justin W. Hart$^{2}$, Peter Stone$^{3}$, and Luis Sentis$^{1,4}$
\thanks{$^{1}$Human Centered Robotics Lab (UT at Austin)}%
\thanks{$^{2}$Department of Computer Science (UT at Austin)}%
\thanks{$^{3}$Department of Computer Engineering (UT at Austin)}%
\thanks{$^{4}$Department of Aerospace Engineering (UT at Austin)}%
}

\begin{document}

\maketitle
\thispagestyle{empty}
\pagestyle{empty}

\begin{abstract}

This paper addresses a novel architecture for person-following robots using active search. The proposed system can be applied in real-time to general mobile robots for learning features of a human, detecting and tracking, and finally navigating towards that person. To succeed at person-following, perception, planning, and robot behavior need to be integrated properly. Toward this end, an active target searching capability, including prediction and navigation toward vantage locations for finding human targets, is proposed. The proposed capability aims at improving the robustness and efficiency for tracking and following people under dynamic conditions such as crowded environments. A multi-modal sensor information approach including fusing an RGB-D sensor and a laser scanner, is pursued to robustly track and identify human targets. Bayesian filtering for keeping track of human and a regression algorithm to predict the trajectory of people are investigated. In order to make the robot autonomous, the proposed framework relies on a behavior-tree structure. Using Toyota Human Support Robot (HSR), real-time experiments demonstrate that the proposed architecture can generate fast, efficient person-following behaviors. 
\end{abstract}

\section{INTRODUCTION}

Following people is a highly desirable skills for mobile robots to support daily chores. To achieve robust and efficient person-following capabilities, perception, robot gaze control, and navigation need to be effectively integrated.

Vision-based human recognition has dramatically improved with new softwares that rely on deep learning based technologies such as YoLo \cite{redmon2016you} and
OpenPose\cite{cao2017realtime}. On the flip side, they have a limited range of sight \cite{morales2012people,basso2013fast}. To resolve this problem, laser-based methods \cite{chung2012detection,kawarazaki2015development},
and various sensor fusion techniques combining face recognition and leg detection have been introduced \cite{scheutz2004fast,fritsch2003multi,bellotto2009multisensor}. However, major difficulties include handling occlusions, identifying target people among crowds, and difficulty on effectively detecting human faces \cite{yuan2015multisensor}. To surpass these limitations this, new techniques have been devised relying on extra features such as the detection of clothes, bags, and shoes \cite{ahmed2015improved}. 
\begin{figure}[t]
 \centering
\includegraphics[width=0.98 \linewidth]{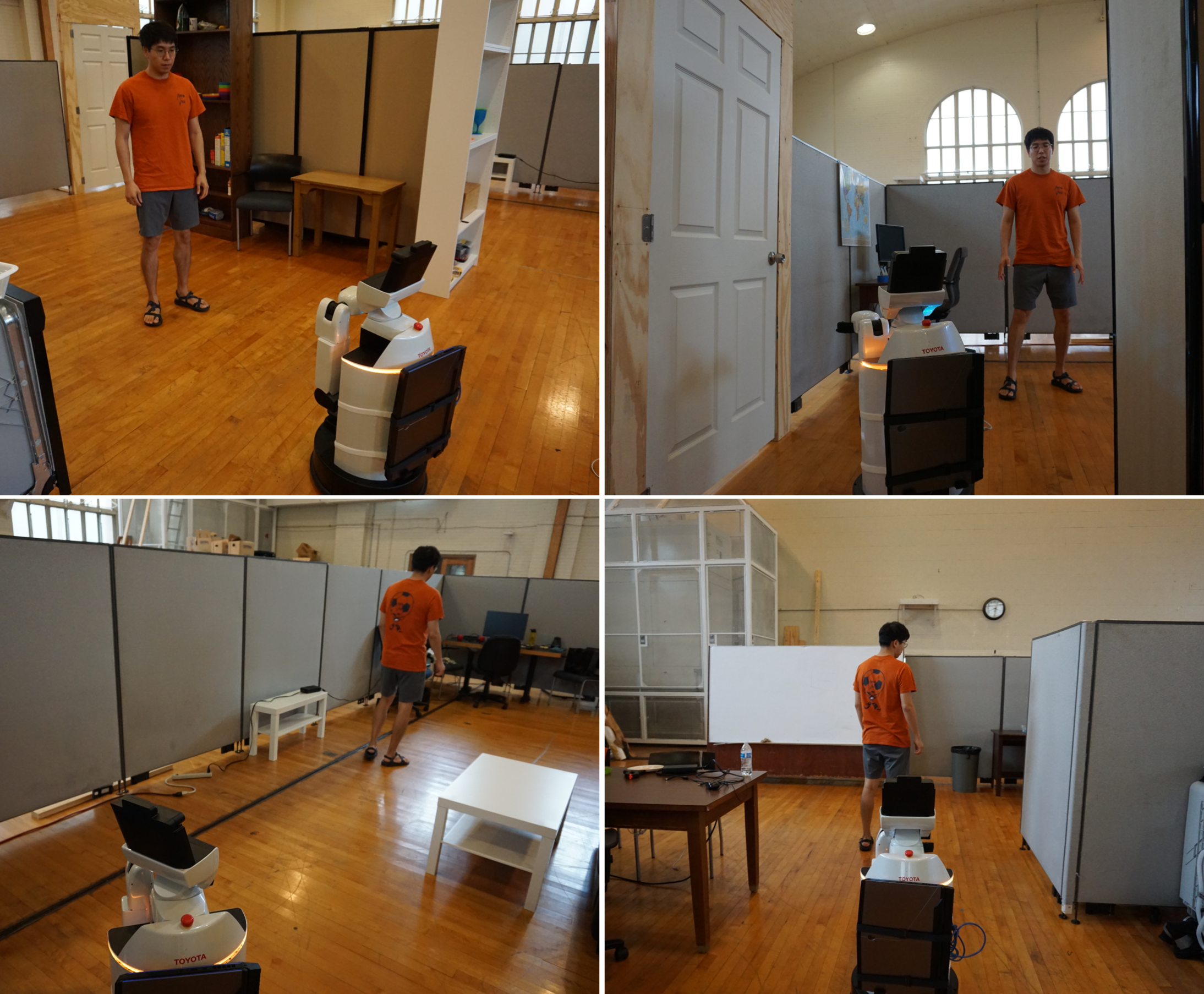}
\protect\caption{Person-following with a mobile robot}
\label{Snapshots}
\end{figure}

Another problem is due to using passive perception techniques where the robot stays stationary thus loosing the target. It is therefore best for robots to achieve active perception such that people can be followed despite their movement \cite{fitzparick2003perception}. We are interested in these questions:
i) where should the robot navigate to? and ii) what should the robot look at? Many researchers have studied this topic withing the topic of active perception or visual sensor planning \cite{chen2011active}. This kind of problem is intractable because there are too many variables. However, using prior knowledge, context, and logical assumptions about the environment it is possible to find solution approximations. If a robot is aware of the connectivity between spaces, when the target suddenly disappears from the view of the robot, one strategy could be to navigate to the last observed location to look for the target. This space connectivity can be simplified by the use of a topology map or graph \cite{portugal2012extracting,thrun2003robotic}. 

Robot skills should be integrated in harmony with the perceptual processes to improve a robot\textquoteright s ability to adapt to the various
dynamic circumstances. For example, actions such as searching for a target, tracking, and navigating should be properly coordinated. To achieve this coordination, a Behavior-Tree framework is applied to sequence the skills \cite{colledanchise2017behavior}.

In that light, the main contribution of this paper is on integrating sensor fusion, context-base motion planning, person movement prediction, and behavior decision making. The rest of paper is organized as follows: Section II gives an overview of the methods to track and follow people. Experimental setup and results are describes in section III. In section IV, we draw conclusions and, finally, we present ideas for future improvements.

\section{FRAMEWORK}

\subsection{Behavior-Tree}

\begin{figure}[t]
    \centering
        {\includegraphics[width=1.0\linewidth]{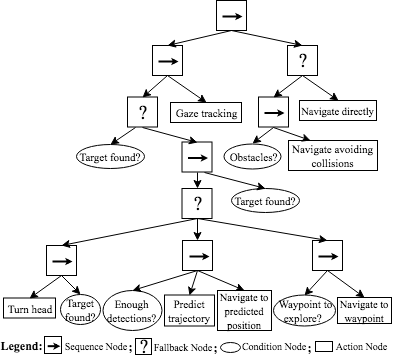}}
    \caption{Behavior-Tree architecture for person-following. }
       \label{Figure:BT_framework}
\end{figure}

A behavior tree (BT) framework \cite{colledanchise2017behavior} allows a robot to achieve autonomous planning in response to various situations. Similarly to (hierarchical) finite state machines, BT's are being utilized because of their modularity, efficiency, and intuitive usage. BT's are considered as generalizations of three classical concepts including Subsumption architecture \cite{brooks1986robust}, Sequential behavior composition \cite{burridge1999sequential}, and Decision trees \cite{nehaniv2002imitation}. BT's contain four types of control flow nodes (fall back, sequence, parallel, and decorator) and two execution nodes (action and condition) \cite{colledanchise2017behavior}. BT's start from a root node toward child nodes with a predefined frequency, and child nodes return one of the values \textit{running}, \textit{success}, or \textit{failure}. Autonomous behaviors of robots can be designed with two fallback nodes and two sequence nodes as shown in Fig.\ref{Figure:BT_framework}. 

\subsection{Map Representation}

 To efficiently describe spatial information of the environment, a polygon-based decomposition method can be used. A map consists of sub regions, while each region can be bounded by a polygon. The connectivity information can be obtained from the map. The main functionality of a high-level map representation is to infer the robot's current location and reason about where it is heading to. This information can be used as an important clue for robots to follow or search for target objects. To compute where a robot is headed, a person-following algorithm is described in Algorithm \ref{waypoint}. The main idea is to compare the relative distance between a robot and each neighborhood and find the location for which relative distance has decreased for a certain duration of time. 

 \begin{algorithm}[t]
 \caption{Way-point\_search()}  \label{waypoint}
 \renewcommand{\algorithmicrequire}{\textbf{Input:}}
 \renewcommand{\algorithmicensure}{\textbf{Output:}}
  \begin{algorithmic}
  \Require {$x$, $g$, $L$, $h_{\tau-T:\tau}$ (robot, graph, map, history)}
  \Ensure {$l^*$ (next way-point)} 
  \State $l_r\gets Get\_current\_location(x)$
  \ForAll{${neighbor}^i\in\mathit{Neighbors(l_r)}$}
  \For{$t=\tau-T$ to $\tau$}
    \State $d^i \gets dist(x,{neighbor}^i)$
    \State $\Delta d^i = d^i_{t+1} -d^i_{t}$
    \State $\sum d^i =\sum d^i+ \Delta d^i$
  \EndFor
  \State $\phi^i = \frac{1}{T} \sum_{\tau-T}^{\tau} \Delta d^i \gets Get\_average\_delta()$ 
  \EndFor
  \State $ \Phi = \cup_i\{ \phi^i \} $ 
  \State $l^{*} = l(i) \gets \argmin_{i}\Phi \gets Get\_heading\_way\-point()$ 
 \While {$m$ is not covered}
  \State $F_c \gets Extract\_Frontier\_Clusters()$
  \ForAll{$f^i\in\mathit{F_c}$}
    \State $c^i\gets$ the closest point in $f^i$ to $l^*$
  \EndFor
  \State $c^{*}\gets \argmax{\bold{U}(c,x,f,l^*)}$
 \EndWhile 
 \Return $c^{*}$
  \end{algorithmic}
 \end{algorithm}

 As for low-level map representations, an occupancy grid mapping scheme, which aims to geometrically represent surroundings with occupancy probabilities, is adopted. This filter uses three types of occupancies including occupied, free, and unknown: $m_k \in \{O(1),F(0),U(0.5)\}$ to characterize a map $m$. From the initial distribution of the occupancy grid which is set to 0.5 for all cells, sequential sensor measurements can update the occupancy grid using inverse sensor models $p_z(m|z)$ for every time step. For the grid within the field of view (FOV), the posterior occupancy probability at time $k+1$, $p_{k+1}(m_{k+1})$ can be obtained via the equation \cite{nuss2016random}: 
 \begin{eqnarray}
  \frac{p_{z_{k+1}}(m_{k+1}|z_{k+1})\cdot p_{k}(m_k)}{p_{z_{k+1}}(m_{k+1}|z_{k+1})\cdot p_{k}(m_k)+p_{z_{k+1}}(\bar{m}_{k+1}|z_{k+1})\cdot p_{k}(\bar{m}_k)} 
  \end{eqnarray}
where $p(\bar{m}) = 1-p(m)$.
  
\subsection{Frontier-Based Exploration}

 Generally, the purpose of exploration for mobile robots is to cover the environment. This exploration strategy can be linked to an uncertainty of the map, described with Shannon's entropy \cite{crupi2016generalized}. Assuming there exist map boundaries, the entropy can be defined as $H(M_t) = -\sum_{i=1}^{N}m^i_t\log(m^i_t)$, where, $m^i_t$ is the $i$-th entry of the map state from the 2D occupancy grid and $N$ denotes the total number of grids. The entropy of map has higher value when there remains many unknown grids. Using this concept, one possible approach is for a robot to choose the best sensing spot that maximizes information gain of the current map. Regarded as the best solution, frontier-based exploration \cite{yamauchi1997frontier} has been widely used, where a frontier reveals the boundaries between known (occupied or free) and unknown areas. This boundaries are potentially informative because they are close to unknown areas. Geometrically, frontiers can be characterized with following equation \cite{jadidi2014exploration}, 
\begin{equation}
 F = ||\nabla \Lambda||_1 - \beta (||\nabla \Lambda_o||_1+ \Lambda_o-0.5)
\end{equation}
where $\nabla$ is the gradient operator, and $\beta$ is a weight factor for the effect of obstacle boundaries.  $||\nabla \Lambda||_1$ defines all boundaries, while the second and third terms meas occupied regions including obstacles and their boundaries. The last constant term, is subtracted to remove the biased probability for unknown region in the occupancy map. the The resulting frontier map $F$ contains only known-free and unknown boundaries and by clustering points in this map we can select goals for further exploration.

 Although this approach was originally developed for SLAM, the core concept of frontiers is quite useful for target search. Frontiers can provide a a practical representation for the selection of the observation location for target search. Therefore, this type of information can be used in combination with a priory knowledge or observation models for the target, to achieve synergistic effects.

\subsection{Utility Function}

A utility function can be defined using frontier clusters. Counting the number of unknown cells in a frontier cluster can be regarded as the amount of information gain as the robot observes that area. In other words, the entropy of clusters is regarded as the number of unknown cells. Therefore, the utility function can be modeled with this information gain. Using this idea, the utility function can be written as:
\begin{equation}
U(c,x,l^*)= \alpha N-{d(x,c)+d(c,l^*)},
\end{equation}
where $d ( \cdot )$ is a distance function and $\alpha$ is the weight factor that affects the information gain and the travel cost. As described in Algorithm \ref{waypoint}, the proposed algorithm will find the best sensing spot to look for the target.  

\subsection{People Detection and Tracking }

In order to be practical, the real-time detection and tracking of people are essential. In our study, RGB-D camera and laser scanners are combined to detect, identify, and track humans. The use of a muti-modal sensor fusion technique can improve the accuracy of recognition and tracking people.
\begin{figure}[t]
    \centering
        {\includegraphics[width=0.98\linewidth]{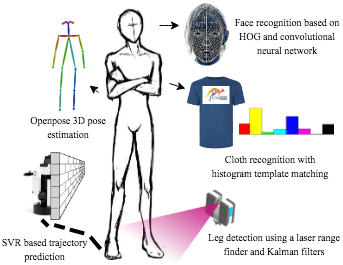}}
    \caption{Person detection and tracking concept}
       \label{Figure:tracking_framework}
\end{figure}

\paragraph{Leg Detection and Tracking}

 Using a laser range finder, human leg patterns can be recognized. we adopt a random forest classifier as described in  \cite{arras2007using}. Each leg position is tracked by an Extended Kalman Filter with a constant velocity assumption. The filter consists of prediction and correction steps. Each filter estimates the state of each leg candidate position and velocity, $x$, $\dot{x}$. The extended Kalman Filter uses a set of linear dynamical systems and a measurement model, $\dot{x}=Ax+Bu+w$, and $z=Hx+v$, respectively. $A$ is a state transition matrix, $B$ is the input matrix, $u$ is input variable, and $w$ is a white Gaussian noise with co-variance $Q$. The measurement variable, $z$, can be modeled with the observation matrix, $H$, and $v$ is the observation noise variable, with co-variance $R$.

  In the case of multi-object tracking, data associations are required. In other words, during the update step, the filter needs to select the best observations to update the current objects being tracked. Here, the Nearest Neighbor based data association method \cite{muja2009fast} is adopted to link new candidates ($i$) and objects ($j$). The main idea is finding pairs $(i,j)$ for all observations and the existing target to minimize the total sum of the distances among all the individual assignments. 

 \paragraph{Human Pose Detection}
  Recently, a real-time convolution neural network based algorithm named OpenPose \cite{wei2016cpm} \cite{cao2017realtime} to estimate 2D human poses with a skeleton tracking has been developed and widely used. A major advantage of this algorithm is that it can robustly detect and track multiple people while providing not only a bounding box of the people but also recognizing the human body parts. It is possible to compute the distance to the person from a robot by using the average coordinates of the recognized body parts, and clustering the point clouds of that average. Therefore, 3D bounding boxes of the human can be obtained. 
  
 \begin{algorithm}[b]
 \caption{Target Identification Strategy}\label{Target Identification Strategy}
 \renewcommand{\algorithmicrequire}{\textbf{Input:}}
 \renewcommand{\algorithmicensure}{\textbf{Output:}}
 \begin{algorithmic}
  \Require {$\mathnormal{B}=\{h_1, h_2, \cdots,h_k\}$ (Current human belief), \\\qquad $k$ : Number of human candidates}
   \Ensure{$True$ \ ($H^*$: Human target) or $False$} 
    \State {\bf Step 1} \textit{Filterwithlegs()}
    \State {\bf Step 2} \textit{Filterwithface()}
    \State {\bf Step 3} \textit{Filterwithclothes()}
 \end{algorithmic}
 \end{algorithm}

\paragraph{Person Identification}

The face recognition package \cite{ageitgey2013} is also applied for our proposed system. It basically uses Histograms of Oriented Gradients (HOG) \cite{dalal2005histograms} to detect faces and face landmark estimation \cite{kazemi2014one} to extract face features. Then, the extracted features are used to train a Deep Convolutional Neural Network to recognize faces. Therefore, human faces can be recognized from the image stream, which leads to identifying people. Because the face is the most definite feature that distinguishes a person from another, the highest level of trust is given to the face recognition process. However, using face information has limitations when the robot is following people. For this, our framework is required to have other clues to identify people such as identifying clothes \cite{lee2008inter}. Thus, firstly, the OpenPose recognition tool is used to extract the region of interest for clothe detection. Then, the color or pattern of that bounding box can be characterized using the histogram intersection algorithm \cite{swain1991color}, which is known for its invariance to translations, scaling, and robustness to occlusions with other objects. Given a learned template ($T$), a histogram intersection is defined as $\sum_{j=1}^n min(I_j,T_j)$, where $n$ is the number of bins. To obtain similarity values between input images $I$ and $T$ it uses the normalized ratio given the following formula, $S=\frac{\sum_{j=1}^n min(I_j,T_j)}{\sum_{j=1}^nT_j}$. Using this histogram-based metric, each image extracted from OpenPose, and the corresponding similarity is computed. Then, a person having the similarity exceeding the critical similarity can be regarded as a target.

\paragraph{Human Belief}

 We start by combining laser-based detection and vision-based detection using Algorithm \ref{Target Identification Strategy}. It is assumed that vision information is more reliable because lasers provide candidates of human legs, rather than precise information. An important point when fusing is that the FOV of the laser sensor and that of the RGB-D camera are different, so that the robot has to be aware of the position of the people being recognized. Similarly to the low-level map representation discussed earlier, the human belief (the probability that human exists in a grid cell) can be expressed using occupancy grid mapping with a Bayesian inference method. Keeping track of the human belief, the robot considers a human to be present in that region until it observes that region. The grid cell inside a current FOV can be iteratively updated over time with following equation, 
\begin{align}
p(h|o) &= \frac{p(o|h)P(o)}{p(o|h)P(h) + P(o|h^c)(1-P(h))}
\end{align}
where $o$ stands for the OpenPose detection results.
\begin{figure*}[t]º
 \centering
\includegraphics[width=0.85\linewidth]{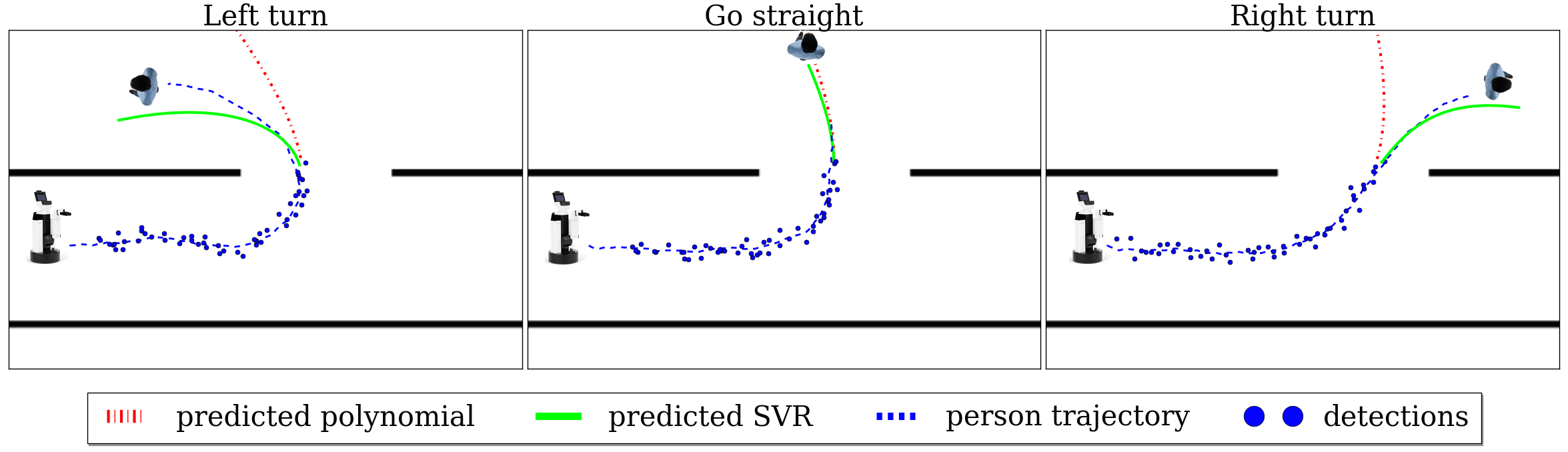}
\protect\caption{Trajectory prediction experiments}
\label{Figure:Prediction_trajectory}
\end{figure*}

\paragraph{Trajectory prediction}

An efficient person-following robot should be able to anticipate where the target might be when it suddenly disappears. The basic idea is to build a regression model from the past history and to extrapolate the person's possible trajectory. A Support Vector Machine Regression (SVR) is adopted to predict the trajectory of people. This can approximate nonlinear relationships, and provide a parsimonious fit, since it relies on kernel functions. Assuming that we have a set of training data where $\boldsymbol{\mathbf{x}}=\left(x_{1},\,x_{2},\,...,\,x_{k}\right)$
is a vector that comprises the input variables ($k$ being the number
of such variables), and that we have $n$ observations: $\left(\mathbf{x}_{1},\,y_{1}\right),\:(\mathbf{x}_{2},\,y_{2}),\,...,\:(\mathbf{x}_{n},\,y_{n})\,,$ where
$y$ is the variable to predict. The problem becomes, according to
the theory of Support Vector Machines (SVM) \cite{Vapnik1998} as that of
finding a function $f(\mathbf{x})=\mathbf{\mathbf{\boldsymbol{\mathbf{\omega}}}}^{T}\phi(\mathbf{x})+b\,,$
and using that function to fit the training data, where: $\boldsymbol{\mathbf{\mathbf{\omega}}}$
is the vector that contains the weights that affect each
predictor; $b$ is a real number; and $\phi$ is a non-linear mapping. SVM theory \cite{Vapnik1999} states that the solution
to this problem is the same as the solution of the equation: 
\begin{equation}
\underset{_{\mathbf{\boldsymbol{\omega}},b,\xi}}{\min}\left(-\mathbf{\boldsymbol{\omega}}^{T}\mathbf{\boldsymbol{\omega}}+C\sum_{i=1}^{n}\xi_{i}\right),\label{eq:6}
\end{equation}
 subject to $y_{i}\left[\mathbf{\boldsymbol{\omega}}^{T}\phi\left(\mathbf{x}_{i}\right)+b\right]\geq1-\xi_{i}\,\,;$$\:\xi_{i}\geq0\,;$$\,\:\left(i=1,\,2,\,...,\,n\right),$
where: $\xi_{i}$ is the error between observed and predicted values,
that is, $\max\left\{ 0,\,\mid y_{i}-f(\mathbf{x}_{i})\mid-\varepsilon\right\} ;$
the parameter $\varepsilon>0$ determines an insensitivity zone around
the fitted model where the error is not taken into account and, $C$
is the penalty parameter that weighs the error in the function that
is minimized. Thus, the term $C\sum_{i=1}^{n}\xi_{i}$ in equation
\ref{eq:6} represents the losses on the training set.

Once $C$, $\varepsilon$ and the parameters of the kernel function
have been selected, this problem has a unique solution. Different
parameters will give different solutions or models. Therefore, the
parameters must be tuned to optimize the model \cite{Garcia-Gonzalo2016}.
The simplest way of performing the parameter tuning is grid search.

\paragraph{Robot control}
 To track the human target, robot gaze control is essential. The gazing behavior is designed to seek for human candidates. For example, when the human target is not visible, a gaze planner forces a robot to look where humans might exist. This information can be obtained from human belief. If the target doesn't exist, the robot will seek for the target using leg candidates. From the geometric relationship between the robot and the target positions, the desired joint values are easily obtained and a PD joint position control is used.
 
Our navigation strategy is to use the $TURN$ and $GO\_STRAIGHT$ commands and the $MOVE\_BASE$ command from the ROS navigation stack \cite{rosnavigation2012}. While the first command is activated when no obstacles exist between the robot and the target, the second command is activated when there exists obstacles in the desired path. The first command is  beneficial: since the laser sensor is the most important observation source, it is effective to track the location of a person in the center of the FOV. Consequently, in order to place a person at the center of the FOV, the first strategy is turning the robot base toward the person and going straight toward the target. 
\begin{figure*}[t]
    \centering
        {\includegraphics[width=1.0\linewidth]{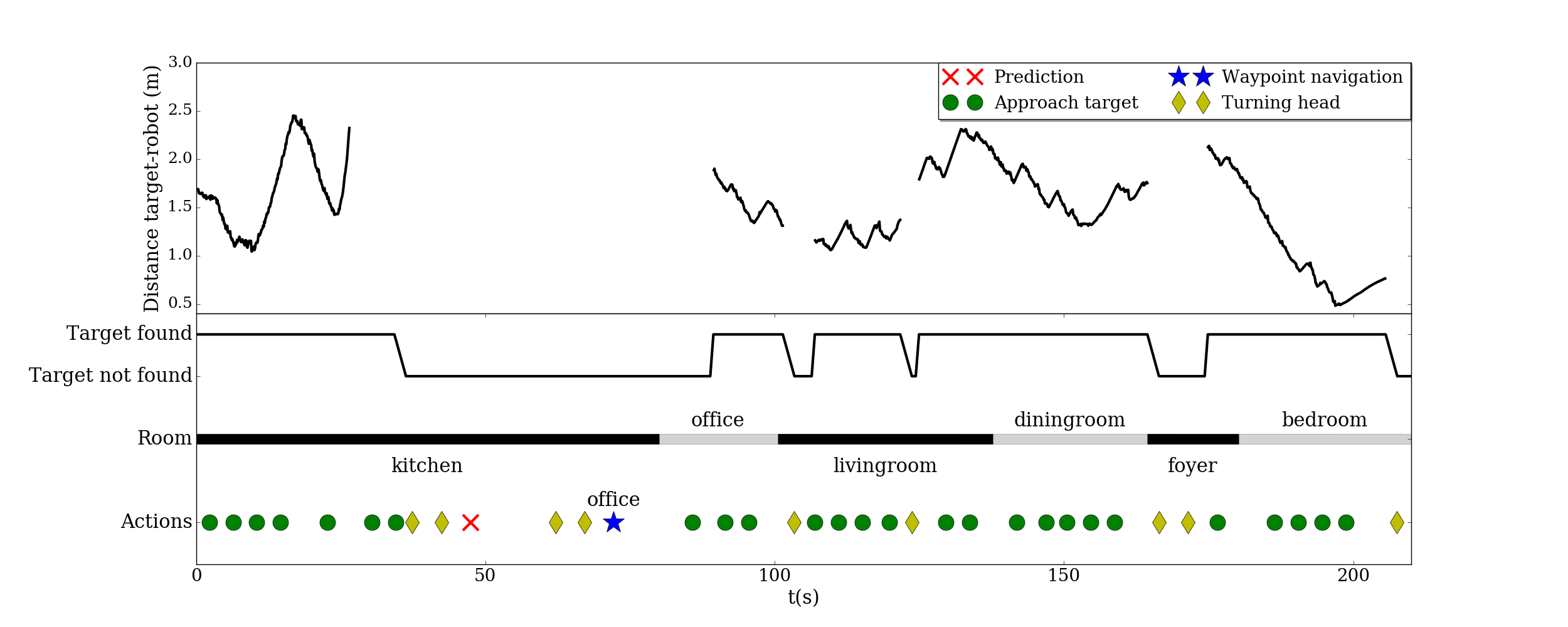}}
      \protect\caption{Information timeline during person-following.}
      \label{Figure:timeline}
\end{figure*}
 
\section{RESULTS}

\subsection{System Description}
 The Toyota Human Support Robot (HSR) is a mobile manipulator \cite{hashimoto2013field} which has been used as a hardware platform for this study. The mobile base of the HSR consists of two omni-wheels and three caster wheels which are located at the front and rear of the robot. The maximum speed of the HSR is approximately $0.22\nicefrac{m}{s}$, maximum step size of the mobile base is $5mm$ and the maximum incline that it can climb is $5^{\circ}$. As for the vision system information, two stereo cameras are mounted around the eyes of the robot, a wide angle camera is on the forehead, a depth camera (Xtion, Asus) is placed on the top of the head to get RGB-D video stream. Furthermore, a laser range scanner, Hokuyo, is mounted at the front bottom the of mobile base platform. The HSR uses two different computers, the main pc is for most sensing and navigation tasks and an Alienware laptop (Intel Core i7-7820HK, GTX 1080) is used for running OpenPose for human detection. All sub-programs for the robot are able to communicate useful information to each other via ROS interfaces. The tested environment is at UT Austin's Anna Hiss Gymnasium (AHG). A prototype of a home-like arena was built to perform tasks including perception, navigation, manipulation, and more. 
 
\subsection{Results}

\subsubsection{Trajectory prediction}
Human walking is quite unpredictable, so that future positions can only be estimated in a limited time and from the previous positions in a short time margin. For that reason, the last observations are the most significant for  trajectory prediction. To address this relative importance in the trajectory estimation, increased weight is given to the last samples. In the case of SVR, the sample weighting re-scales the $C$ parameter, which means that the model puts more emphasis on getting these points right \cite{Han2014}. To compare the experimental results, a three-degree polynomial regression is also implemented.

The parameters of SVR with radial basis function (RBF) kernel were selected while the grid search method and the optimal values that we have used are $C=1000.0$, $\varepsilon=0.01$, and $\gamma=1.0$. The prediction algorithms were tested with three difficult, but common situations regarding person-following: when the target goes through a door, she hides completely from the robot and can turn left, go straight or turn right. The results obtained are shown in figure \ref{Figure:Prediction_trajectory}. In the case of turning right or left, the polynomial regression prediction diverges from the real trajectory, while the SVR prediction gets a good approximation. However, we must take into account the limitations of the prediction algorithm since the extrapolation process is full of uncertainties and can produce meaningless predictions. The divergence with the actual trajectory is a characteristic of extrapolation methods. Thus, estimations might only be considered valid within a limited time range. 
\begin{figure}[t]
 \centering
  \includegraphics[width=0.99\linewidth]{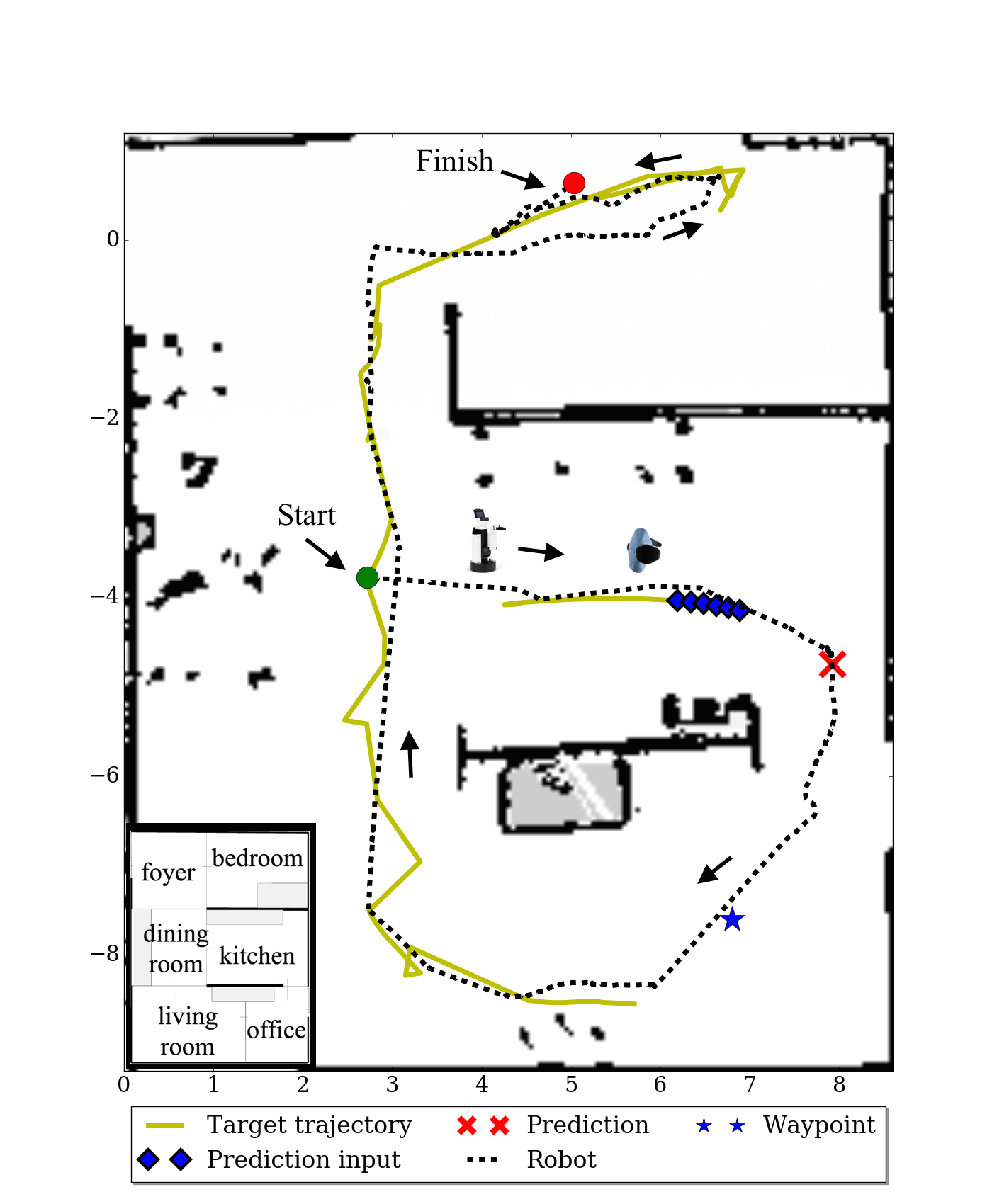}
     \protect\caption{Trajectories projected on the house map with map annotations.}
        \label{Figure:trajectory}
\end{figure}

\subsubsection{Person-Following Performance}

 We address the case of person-following in indoor environments to validate performance of the proposed architecture. The result is shown in Fig. \ref{Figure:timeline} and Fig.\ref{Figure:trajectory}. After learning the target's face and clothing information, robot following is initiated from a starting point. At a certain moment, the target was lost during its way from the kitchen to the office. The first strategy of the robot is to try to predict where person has gone via SVR-based prediction using the input data. Then, the robot decided to go to that location to look for the existence of the person. Since it failed to seek the target using the robot's gaze for that position, the way-point search is activated for further search. Using the Algorithm \ref{waypoint}, the robot navigated to the office location. There, the robot re-identified the target and started to follow her again. The robot status of awareness of target and corresponding actions  during the experiment are well described in the second and third rows in Fig.\ref{Figure:timeline}.
 
 Three main achievements should be highlighted. Firstly, the versatility of implementing the person-following architecture with the support of a behavior tree. As it can be seen in the action sequence it confers the robot the ability to perform complex tasks based on reasoning about simple tasks. Secondly, regarding person tracking, Fig. \ref{Figure:timeline} shows that the robot is able to track, follow and successfully re-identify the target in a dynamic environment. Finally, the robustness of the architecture has been achieved by using active search techniques such as the SVR based trajectory prediction and the way-point search that allowed the robot to re-identify the target and complete its task successfully.
 
\section{CONCLUSIONS}

 This paper proposes a novel architecture for a human-following robot that allows robust and efficient tracking in highly dynamic environments. Autonomous behavior planning can successfully coordinate perception and navigation for the robot. Information regarding regions on the map and frontiers from the current observations are used to obtain best observational location. A trajectory prediction algorithm based on Support Vector Machine (SVM) has also been implemented. Finally, we achieved robust autonomous person-following using the HSR.

Although the proposed framework is effective, there are still a lot of situations that can cause the robot to freeze or fail. Similarly to humans, to cope with uncertain situations, reinforcement learning can improve the performance. A probabilistic model can be developed that contributes to solve the uncertainties that arise. On the other hand, regarding following behaviors, the utility function or the cost must be formulated mathematically. What should be the cost function for person-following? These kind of criteria could be learned through exploration of human-robot interactions. 

\bibliographystyle{IEEEtran}
\bibliography{icra2018}

\begin{thebibliography}{10}
\providecommand{\url}[1]{#1}
\csname url@samestyle\endcsname
\providecommand{\newblock}{\relax}
\providecommand{\bibinfo}[2]{#2}
\providecommand{\BIBentrySTDinterwordspacing}{\spaceskip=0pt\relax}
\providecommand{\BIBentryALTinterwordstretchfactor}{4}
\providecommand{\BIBentryALTinterwordspacing}{\spaceskip=\fontdimen2\font plus
\BIBentryALTinterwordstretchfactor\fontdimen3\font minus
  \fontdimen4\font\relax}
\providecommand{\BIBforeignlanguage}[2]{{%
\expandafter\ifx\csname l@#1\endcsname\relax
\typeout{** WARNING: IEEEtran.bst: No hyphenation pattern has been}%
\typeout{** loaded for the language `#1'. Using the pattern for}%
\typeout{** the default language instead.}%
\else
\language=\csname l@#1\endcsname
\fi
#2}}
\providecommand{\BIBdecl}{\relax}
\BIBdecl

\bibitem{redmon2016you}
J.~Redmon, S.~Divvala, R.~Girshick, and A.~Farhadi, ``You only look once:
  Unified, real-time object detection,'' in \emph{Proceedings of the IEEE
  conference on computer vision and pattern recognition}, 2016, pp. 779--788.

\bibitem{cao2017realtime}
Z.~Cao, T.~Simon, S.-E. Wei, and Y.~Sheikh, ``Realtime multi-person 2d pose
  estimation using part affinity fields,'' in \emph{CVPR}, 2017.

\bibitem{morales2012people}
L.~Y. Morales~Saiki, S.~Satake, R.~Huq, D.~Glas, T.~Kanda, and N.~Hagita, ``How
  do people walk side-by-side?: using a computational model of human behavior
  for a social robot,'' in \emph{Proceedings of the seventh annual ACM/IEEE
  international conference on Human-Robot Interaction}.\hskip 1em plus 0.5em
  minus 0.4em\relax ACM, 2012, pp. 301--308.

\bibitem{basso2013fast}
F.~Basso, M.~Munaro, S.~Michieletto, E.~Pagello, and E.~Menegatti, ``{Fast and
  robust multi-people tracking from RGB-D data for a mobile robot},'' in
  \emph{Intelligent Autonomous Systems 12}.\hskip 1em plus 0.5em minus
  0.4em\relax Springer, 2013, pp. 265--276.

\bibitem{chung2012detection}
W.~Chung, H.~Kim, Y.~Yoo, C.-B. Moon, and J.~Park, ``The detection and
  following of human legs through inductive approaches for a mobile robot with
  a single laser range finder,'' \emph{IEEE transactions on industrial
  electronics}, vol.~59, no.~8, pp. 3156--3166, 2012.

\bibitem{kawarazaki2015development}
N.~Kawarazaki, L.~T. Kuwae, and T.~Yoshidome, ``Development of human following
  mobile robot system using laser range scanner,'' \emph{Procedia Computer
  Science}, vol.~76, pp. 455--460, 2015.

\bibitem{scheutz2004fast}
M.~Scheutz, J.~McRaven, and G.~Cserey, ``Fast, reliable, adaptive, bimodal
  people tracking for indoor environments,'' in \emph{Intelligent Robots and
  Systems, 2004.(IROS 2004). Proceedings. 2004 IEEE/RSJ International
  Conference on}, vol.~2.\hskip 1em plus 0.5em minus 0.4em\relax IEEE, 2004,
  pp. 1347--1352.

\bibitem{fritsch2003multi}
J.~Fritsch, M.~Kleinehagenbrock, S.~Lang, T.~Pl{\"o}tz, G.~A. Fink, and
  G.~Sagerer, ``Multi-modal anchoring for human-robot interaction,''
  \emph{Robotics and Autonomous Systems}, vol.~43, no.~2, pp. 133--147, 2003.

\bibitem{bellotto2009multisensor}
N.~Bellotto and H.~Hu, ``Multisensor-based human detection and tracking for
  mobile service robots,'' \emph{IEEE Transactions on Systems, Man, and
  Cybernetics, Part B (Cybernetics)}, vol.~39, no.~1, pp. 167--181, 2009.

\bibitem{yuan2015multisensor}
J.~Yuan, H.~Chen, F.~Sun, and Y.~Huang, ``Multisensor information fusion for
  people tracking with a mobile robot: A particle filtering approach.''
  \emph{IEEE Trans. Instrumentation and Measurement}, vol.~64, no.~9, pp.
  2427--2442, 2015.

\bibitem{ahmed2015improved}
E.~Ahmed, M.~Jones, and T.~K. Marks, ``An improved deep learning architecture
  for person re-identification,'' in \emph{Proceedings of the IEEE Conference
  on Computer Vision and Pattern Recognition}, 2015, pp. 3908--3916.

\bibitem{fitzparick2003perception}
P.~Fitzpatrick, ``Perception and perspective in robotics,'' in
  \emph{Proceedings of the Annual Meeting of the Cognitive Science Society},
  2003.

\bibitem{chen2011active}
S.~Chen, Y.~Li, and N.~M. Kwok, ``Active vision in robotic systems: A survey of
  recent developments,'' \emph{The International Journal of Robotics Research},
  vol.~11, no.~30, pp. 1343--1377, 2011.

\bibitem{portugal2012extracting}
D.~Portugal and R.~Rocha, ``Extracting topological information from grid maps
  for robot navigation,'' in \emph{ICAART 2012 - Proceedings of the 4th
  International Conference on Agents and Artificial Intelligence - INSTICC}.

\bibitem{thrun2003robotic}
\BIBentryALTinterwordspacing
S.~Thrun, ``Exploring artificial intelligence in the new millennium,''
  G.~Lakemeyer and B.~Nebel, Eds.\hskip 1em plus 0.5em minus 0.4em\relax San
  Francisco, CA, USA: Morgan Kaufmann Publishers Inc., 2003, ch. Robotic
  Mapping: A Survey, pp. 1--35. [Online]. Available:
  \url{http://dl.acm.org/citation.cfm?id=779343.779345}
\BIBentrySTDinterwordspacing

\bibitem{colledanchise2017behavior}
M.~Colledanchise, ``Behavior trees in robotics,'' Ph.D. dissertation, KTH Royal
  Institute of Technology, 2017.

\bibitem{brooks1986robust}
R.~Brooks, ``A robust layered control system for a mobile robot,'' \emph{IEEE
  journal on robotics and automation}, vol.~2, no.~1, pp. 14--23, 1986.

\bibitem{burridge1999sequential}
R.~R. Burridge, A.~A. Rizzi, and D.~E. Koditschek, ``Sequential composition of
  dynamically dexterous robot behaviors,'' \emph{The International Journal of
  Robotics Research}, vol.~18, no.~6, pp. 534--555, 1999.

\bibitem{nehaniv2002imitation}
C.~L. Nehaniv and K.~Dautenhahn, \emph{Imitation in animals and
  artifacts}.\hskip 1em plus 0.5em minus 0.4em\relax MIT press, 2002.

\bibitem{nuss2016random}
D.~Nuss, S.~Reuter, M.~Thom, T.~Yuan, G.~Krehl, M.~Maile, A.~Gern, and
  K.~Dietmayer, ``A random finite set approach for dynamic occupancy grid maps
  with real-time application,'' \emph{arXiv preprint arXiv:1605.02406}, 2016.

\bibitem{crupi2016generalized}
V.~Crupi, J.~D. Nelson, B.~Meder, G.~Cevolani, and K.~Tentori, ``Generalized
  information theory meets human cognition: Introducing a unified framework to
  model uncertainty and information search,'' \emph{Cognitive science}, 2016.

\bibitem{yamauchi1997frontier}
B.~Yamauchi, ``A frontier-based approach for autonomous exploration,'' in
  \emph{Computational Intelligence in Robotics and Automation, 1997. CIRA'97.,
  Proceedings., 1997 IEEE International Symposium on}.\hskip 1em plus 0.5em
  minus 0.4em\relax IEEE, 1997, pp. 146--151.

\bibitem{jadidi2014exploration}
M.~G. Jadidi, J.~V. Mir{\'o}, R.~Valencia, and J.~Andrade-Cetto, ``Exploration
  on continuous gaussian process frontier maps,'' in \emph{Robotics and
  Automation (ICRA), 2014 IEEE International Conference on}.\hskip 1em plus
  0.5em minus 0.4em\relax IEEE, 2014, pp. 6077--6082.

\bibitem{arras2007using}
K.~O. Arras, O.~Mart{\'i}nez-Mozos, and W.~Burgard, ``{Using boosted features
  for the detection of people in 2D range data},'' in \emph{Robotics and
  Automation, 2007 IEEE International Conference on}.\hskip 1em plus 0.5em
  minus 0.4em\relax IEEE, 2007, pp. 3402--3407.

\bibitem{muja2009fast}
M.~Muja and D.~G. Lowe, ``Fast approximate nearest neighbors with automatic
  algorithm configuration.'' \emph{VISAPP (1)}, vol.~2, no. 331-340, 2009.

\bibitem{wei2016cpm}
S.-E. Wei, V.~Ramakrishna, T.~Kanade, and Y.~Sheikh, ``Convolutional pose
  machines,'' in \emph{CVPR}, 2016.

\bibitem{ageitgey2013}
\BIBentryALTinterwordspacing
A.~Geitgey. Face recognition package. [Online]. Available:
  \url{https://github.com/ageitgey/face_recognition}
\BIBentrySTDinterwordspacing

\bibitem{dalal2005histograms}
N.~Dalal and B.~Triggs, ``Histograms of oriented gradients for human
  detection,'' in \emph{Computer Vision and Pattern Recognition, 2005. CVPR
  2005. IEEE Computer Society Conference on}, vol.~1.\hskip 1em plus 0.5em
  minus 0.4em\relax IEEE, 2005, pp. 886--893.

\bibitem{kazemi2014one}
V.~Kazemi and J.~Sullivan, ``One millisecond face alignment with an ensemble of
  regression trees,'' in \emph{Proceedings of the IEEE Conference on Computer
  Vision and Pattern Recognition}, 2014, pp. 1867--1874.

\bibitem{lee2008inter}
J.~Lee, W.~B. Knox, and P.~Stone, ``Inter-classifier feedback for human-robot
  interaction in a domestic setting,'' 2008.

\bibitem{swain1991color}
M.~J. Swain and D.~H. Ballard, ``Color indexing,'' \emph{International journal
  of computer vision}, vol.~7, no.~1, pp. 11--32, 1991.

\bibitem{Vapnik1998}
V.~Vapnik, \emph{Statistical Learning Theory}.\hskip 1em plus 0.5em minus
  0.4em\relax New York: Wiley-Interscience, 1998.

\bibitem{Vapnik1999}
------, ``An overview of statistical learning theory,'' \emph{IEEE T. Neural
  Networks}, vol.~5, no.~10, pp. 988--999, 1999.

\bibitem{Garcia-Gonzalo2016}
{E. Garc{\'\i}a-Gonzalo, A.J.A. Santos, J. Mart{\'\i}nez-Torres, H. Pereira, R.
  Simoes, P.J. Garc{\'\i}a-Nieto, and O. Anjos}, ``{Prediction of five softwood
  paper properties from its density using SVR},'' \emph{BioResources}, vol.~11,
  no.~1, pp. 1892--1904, 2016.

\bibitem{rosnavigation2012}
\BIBentryALTinterwordspacing
E.~Marder-Eppstein. {ROS} navigation stack. [Online]. Available:
  \url{www.ros.org/wiki/navigation}
\BIBentrySTDinterwordspacing

\bibitem{hashimoto2013field}
K.~Hashimoto, F.~Saito, T.~Yamamoto, and K.~Ikeda, ``A field study of the human
  support robot in the home environment,'' in \emph{Advanced Robotics and its
  Social Impacts (ARSO), 2013 IEEE Workshop on}.\hskip 1em plus 0.5em minus
  0.4em\relax IEEE, 2013, pp. 143--150.

\bibitem{Han2014}
X.~Han and L.~Clemmensen, ``On weighted support vector regression,''
  \emph{Quality and Reliability Engineering}, no.~10, 2014.

\end{thebibliography}

\end{document}